1# Feasibility Assessment of a Cost-Effective Two-Wheel Kian-I Mobile Robot for Autonomous Navigation

Amin Abbasi, Somaiyeh MahmoudZadeh, Amirmehdi Yazdani, Ata Jahangir Moshayedi*Abstract*— A two-wheeled mobile robot, namely Kian-I, is designed and prototyped in this research. The Kian-I is comparable with Khepera-IV in terms of dimensional specifications, mounted sensors, and performance capabilities and can be used for educational purposes and cost-effective experimental tests. A motion control architecture is designed for Kian-I in this study to facilitate accurate navigation for the robot in an immersive environment. The implemented control structure consists of two main components of the path recommender system and trajectory tracking controller. Given partial knowledge about the operation field, the path recommender system adopts B-spline curves and Particle Swarm Optimization (PSO) algorithm to determine a collision-free path curve with translational velocity constraint. The provided optimal reference path feeds into the trajectory tracking controller enabling Kian-I to navigate autonomously in the operating field. The trajectory tracking module eliminate the error between the desired path and the followed trajectory through controlling the wheels' velocity. To assess the feasibility of the proposed control architecture, the performance of Kian-I robot in autonomous navigation from any arbitrary initial pose to a target of interest is evaluated through numerous simulation and experimental studies. The experimental results demonstrate the functional capacities and performance of the prototyped robot to be used as a benchmark for investigation and verification of various mobile robot algorithms in the laboratory environment.

*Index Terms*— Kian-I mobile robot, Path Recommender System, Trajectory Tracking, Autonomy, Motion controller## 1. Introduction

Two-wheeled robots have diverse applications due to their high versatility, mobility, and agility in both indoor and outdoor environments. An autonomous robot should be capable of positioning, localization, and autonomous navigation with minimum position calibration error to mitigate the challenges of real-world experiments, which is known as the navigation autonomy. The level of autonomy in mobiles robots is a direct function of hardware specifications including geometrical design, navigation aids, and computational power in conjunction with the software specifications that are mission and motion planning architectures and their functionalities.

In this research, an affordable-cost two-wheeled mobile robot, namely Kian-I, is designed, developed, and prototyped. The aim is to benefit form a functional and cost-effective mobile robot as a benchmark for educational purposes in particular for investigation and verification of different mobile robot algorithms experimentally. To this end, different phases, including the physical design, system integration, sensing, and control architecture, are developed and presented in detail. A motion control architecture is designed for Kian-I in this study to facilitate accurate navigation of the robot in immersive environments. The proposed control architecture consists of two main components of the path recommender system and trajectory tracking controller incorporated knowledge of the state-of-the art in prototyping mission and motion planning algorithms; examples of these are described briefly in the following paragraphs.

A modular chain-wheeled mobile robot with the focus on teaching and research application was prototyped in [1]. The standard Robotic Operating System (ROS) platforms were used for the control, sensing, and communication of the robot. The modular physical design of the robot considered mounting and implementation of different sensors and actuators enabling the robot to handle various operation tasks. In [2], a wheeled-mobile robot based on the Arduino Uno R3 microcontroller was developed. The robot was equipped with servomotor-driven wheels, Bluetooth wireless communication, as well as different object-detection sensors to avoid obstacles. However, the functionality of the robot was limited only to perform very basic actions such as moving forward/ backward, turning left/right, and stopping movement. In [3], a delivery robot platform with 6-DOF manipulator, called HuboQ was designed to handle a human transporter task. A four-wheel Zero-Moment Point (ZMP) control configuration was for self-balancing and human-riding made this robot suitable for transportation, inspection, and monitoring in industrial sectors. In [4], a paradigmatic mobile robot framework was designed to benefit from an Inspection Robotic System (IRS) in practice for monitoring the industrial infrastructure. The developed platform enabled the robot to monitor an industrial site using different inspection technics, such as thermal image capturing and 3D scan of mechanical components. The engineering feasibility of the model and interoperability of the mobile robot in immersive environments was experimentally tested and evaluated. The results of the tests showed the effectiveness and functionality of the design in practice. In another work [5], an Omni-directional wheeled robot was designed to establish material conveying tasks in an industrial environment, where the robot was equipped with multiple perceptions and localization devices, such as visual sensors and wheel encoders. A particular type of wheel, called Mecanum, used in [5] that provided flexible maneuverability for the platform. In [6], an Omni-directional

---

Amin Abbasi, Department of Electrical Engineering, Azad University of Khoemeinishar, Esfahan, Iran (e-mail: AminAbbasi.res@gmail.com).

Somaiyeh MahmoudZadeh, School of IT, Deakin University, Geelong, VIC 3220, Australia (e-mail: S.MahmoudZadeh@deakin.edu.au).

Amirmehdi Yazdani, College of Science, Health, Engineering and Education, Murdoch University, Perth, WA 6150, Australia (e-mail: Amirmehdi.Yazdani@murdoch.edu.au).

Ata Jahangir Moshayedi, School Of Information Engineering, Jiangxi University Of Science and Technology, Jiangxi, 341000, China (e-mail: ajm@jxust.edu.cn).1

mobile robot for RoboCup competitions was designed. The control architecture of the proposed robot was based on the Arduino Mega microcontroller and the robot was capable of distinguishing the color of the ball in the playground using on-board cameras and controlling its wheels' velocity toward the ball.

The motion planning of a mobile robot is provided by simultaneous working of three substantial modules called guidance, navigation, and control which are collectively known as GNC architecture. The navigation module is responsible for perception and localization based on the information collected by the sensors. The guidance module, or path planning refers to that activity where the robot perceives the surrounding environment based on sensory information and then decides to autonomously plan a feasible/optimal path from the initial location toward the destination. Finally, the control module known as trajectory tracking, is responsible to produce control inputs so that the generated path is tracked with minimum error. There are numerous studies in the state-of-the art with focus on design and development of GNC framework for mobile robots. For instance, a humanoid wheeled robotic platform was employed by Lee et al., (2020) for fetching and serving the drink in both static indoor and dynamic outdoor environments comprising the ability of object detection, path planning, grasping, localization, navigation and motion control [7]. In this research [7], A* method was utilized for trajectory planning, and the performance of the robot was implemented in "MoveIt" software platform. A non-holonomic A* optimizer was also used by Zhang et al. in (2019) for path planning of a car-like vehicle in a highly uneven ground, where the surface geometric information was obtained by a 3D light detection sensor, and the suspension-based traversability calculation used to enable the robot moving on a rough terrain surfaces [8]. To apply A* as a grid-search-based algorithm on path planning problem, first the search space should be decomposed to a graph feature (or grid environment) to specify the collision boundaries and forbidden sections, and then the path is produced by patching the free sections based on defined cost function. The grid-search-based methods are criticized due to their discrete state transitions, which restrict the robot's motion flexibility to limited set of directions. The grid-based algorithms are also inefficient in extremely large or complex workspaces, where growth of the cell numbers make data rendering intractable.

A modified multi-objective PSO was used to solve a path planning problem for car-like robots under the limited length and the terrain robustness constraints in a rough environment [9]. The modeling of the rough terrain environment was created using a contour map, height map, and grayscale map. The proposed algorithm was tested in two simulation platforms of Microsoft Robotic Developer Studio 4 and MATLAB. However, this research is limited to simulation and lack experimental validation in a real-world environment.

In [10], a path planning algorithm based on deep reinforcement learning for a wheeled mobile robot was proposed. In this study, the Double Deep Q-Network (DDQN) was modified by an optimized tree structure to perform an effective path planning. The suggested path planner was implemented by ROS platform, and the simulation results revealed the efficiency of the proposed method compared to conventional DDQN in dealing with complex slope ground; however, no experimental validation is performed to assess the algorithm's capability in a real-world environment. A path recommender system using an improved version of DDQN algorithm to control the overestimation problem associated with the conventional DDQN method was developed in [11]. The authors conducted a relatively gentle operation based on an improved ε-greedy strategy to use the previous experiences in updating the target value, to reduce the error between the estimated value and the real value, and to diminish the overestimation impact on the path planning process in a grid environment. However, this study also remains in simulation environment while the vehicles kinematic and motion dynamic is not clearly specified.

On the other hand, there exists a plethora of documented research on trajectory tracking algorithms of mobile robots. A trajectory tracking controller was designed for a specific type of mobile robots, named Hilare, where the motion control of the robot was carried out by controlling the velocity of its stepper-motor-driven wheels [12]. The adaptive control parameters in [12] were optimized by multi-objective PSO to minimize the trajectory and velocity tracking errors. The produced results show the trajectory tracking controller designed in this research was able to follow the given paths with a reasonable error. In [13] a second-order sliding mode controller was designed for a four-wheeled skid-steered mobile robot to track a predefined trajectory under external disturbance and parametric uncertainties. The control inputs (angular and transitional velocity) of the robot were fed to a real robot, and the experimental performance validated the simulation results. However, the evaluation of the proposed controller for tracking more difficult reference trajectories, such as spiral and 8-shape curves, was remained unaddressed. Liu and Wang (2019) implemented a local trajectory planning approach to generate a reasonable trajectory for a ground service robot, while the obstacle avoidance and the trajectory curvature limit have been considered as the main constraints of the problem [14]. This study evaluated the proposed idea in MATLAB simulation environment, while the experimental validation of the performance of the proposed method in the real world environment remains unaddressed.

In [15], a sliding mode control model using PID sliding surface was presented for trajectory and velocity tracking of a differentially-driven wheeled robot. The stability of the proposed controller was proven by Lyapunov stability and the system was able to navigate the robot in the presence of uncertainties and disturbances. In [16], an optimization-based control model established on the nonlinear control theory and integral feedback technique was proposed for trajectory tracking of a differentially-driven wheeled mobile robot. Compared to the previous research [15], the controller proposed in [16] has a better capability in dealing with nonlinear behavior of the robot's kinematic and dynamic model and demonstrated better robustness under uncertainties and disturbance.

Zhang and Liu [17] suggested a fractional-order PD controller for trajectory tracking of a wheeled mobile robot. In this study [17], a robust controller tuning specification is employed for satisfying the flatness of the phase curve in a frequency interval to improve the overall controlled system robustness, and the proposed algorithm is validated experimentally. In [18], a dynamic trajectory tracking controller for non-holonomic movement of a wheeled industrial manipulator equipped with rotary-sliding (R-S) joints is introduced. The authors conducted a recursive predictive control model to systematically find the kinematic control rules, while Gibbs-Appell (G-A) technique is applied to model the dynamics of the manipulator and mitigate the difficulties of Lagrange Multipliers resulted from non-holonomic constraints. The emphasized approach in this research is capable of dynamic modelling of the platform motions and underneath trajectory control complexities and limitations of wheeled industrial manipulators.

This research deals with the design and development of a cost-effective prototype of a two-wheeled mobile robotic platform, namely Kian-I, that can be utilized in the field of robotic science and technology for both research and educational activities. Different prototyping phases, including the design, system integration, sensing and control modulation, simulation, development, and experimental evaluation are thoroughly presented in this paper. The development of the robot's motion control unit comprising two main components of the path recommender system and trajectory tracking controller, facilitating accurate navigation of the robot in immersive environments, is also presented. By this design, a considerable level of autonomy is achieved upon a multi-layered and deliberative control architecture, where the robot performs on-board decision-making without reliance on the human supervisor. These enable Kian-I, for localization on the environment map and autonomous collision-free path generation and maneuver toward the target of interest. To investigate the feasibility, reliability, and redundancy of the proposed mobile platform in practice, extensive simulation and experimental tests are carried out. These include:

- Functional and efficiency assessment of the physical structure design in terms of sizing, modularity and payload handling in compare with a standard off-the-shelf mobile robot platform such Khepera-IV;
- Deployment procedure, compatibility verification, performance evaluation, and robustness assessment of the proposed control architecture on Kian-I platform in various workspaces.

The rest of this paper is organized as follows. Section 2 introduces the hardware characteristics, kinematic and dynamical model of the Kian-I robot and compares the key features of this robot with Khepera-IV. The system architecture including the path recommender system and the trajectory tracking controller is presented in section 3. In section 4, the performance of the control architecture is investigated through extensive simulation and experimental studies. Section 5 concludes the achievements of this research.

## 2. Physical Design and Development

Figure 1 illustrates a 3D sketch of the Kian-I where Fig.1 (a) shows the designed robot structure and Fig.1 (b) represents the robot covered by a polyethylene skin. The main chassis and mechanical frame are designed using CATIA software. By design, some geometrical features and main capabilities of Khepera-IV have been considered for Kian-I. Table1 compares the technical specifications of the Khepera-IV and Kian-I. The 3D design encapsulates spaces for different sensor suites on the robot platform. For example, three ultrasonic sensors with angle of 90 degrees are located on the robot. The camera placed on front of the robot and four infrared sensors are located under the chassis for providing line tracking and collision avoidance tasks. The robot is supported by wireless communications for sending and receiving data and images through radio frequency modules.

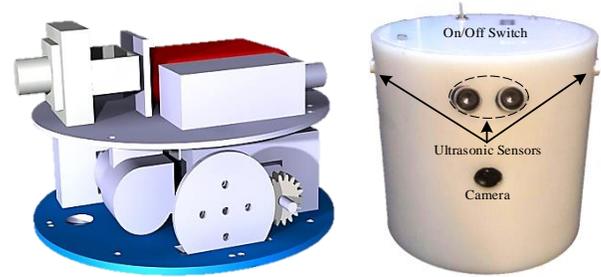

Fig.1. (a) Robot Chassis design (b) The final look of the robot

**Table 1.** Hardware specification of Khepera-IV and designed robot Kian-I

| Hardware | Khepera-IV | Kian-I |
|---|---|---|
| DC motors | DC Motors 1.96w | NAMIKI 120RPM 12V |
| Ultrasonic | 400PT12B | SRF05 |
| Infrared | TCRT5000 | TCRT5000 |
| Battery | Li-Po 7.4v 3400mAh | Li-Lo 11.7v 4400mAh |
| Memory | Micro SD RAM | Micro SD RAM |
| IMU | LSM330DLC | GY801 |
| Wireless Communication | Embedded | NRF24L01 |
| Image Transmitter | Bluetooth | TS351 |
| Image Receiver | Bluetooth | RC805 |
| Microcontroller | Gumstix Overo | Arduino MEGA |
| DC Motor Driver | Embedded | IC - L298N |
| Camera | MT9V034C12ST | FPV700TVL |

The data login and communication interface between the electronic and electrical parts of the robot, including the microcontroller, embedded sensors, and DC motors, is provided by a printed circuit board (PCB) designed by PROTEUS software. Both sides of the PCB are shown in Fig.2(a). The PCB is located on top of the main chassis. The assembled robot is depicted in Fig.2(b).

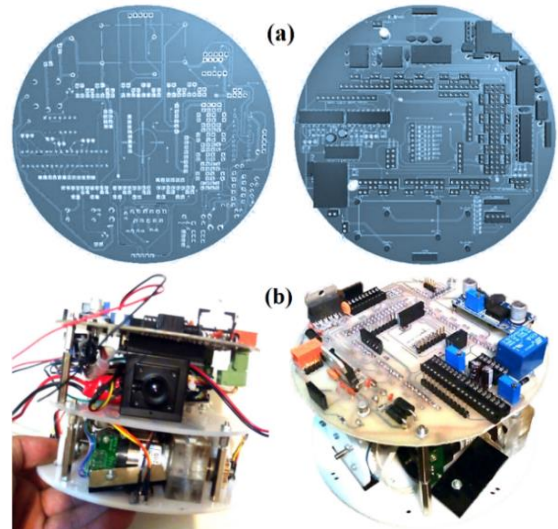

Fig.2 (a) The PCB schematics of the robot; (b) The assembled robot.

### 2.1 Robot Kinematic and Motion Dynamic Model

Similar to Khepera-IV [19], Kian-I also operates by independent speed control of each wheel, while two other spin wheels provide the robot with the ability of rotating in all directions. The DC motors are used to drive the robot and the outputs of the control system are the voltage applied to the left and right wheels. Fig. 3 shows a motion schematic on the Kian-I robot.

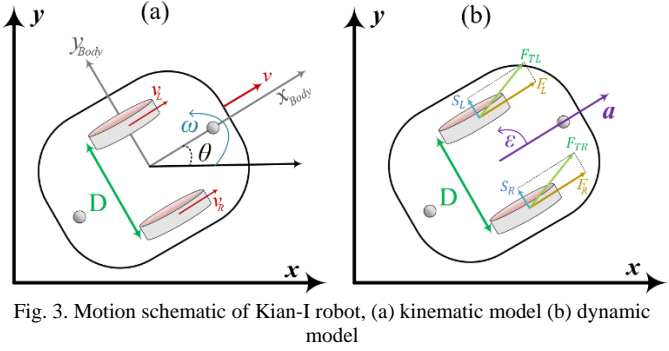

Fig. 3. Motion schematic of Kian-I robot, (a) kinematic model (b) dynamic model

As shown in Fig. 3 (a), $v$ and $\omega$ are the translational and angular velocities of the robot, respectively; $v_R$ and $v_L$ correspond to tangential velocity on the right and left wheel speeds, while $\omega_L$ and $\omega_R$ are the right and left angular velocities, respectively. $D$ is the distance between two wheels and the robot yaw angle is also presented by $\theta$. The robot's motion is controlled by the translational angular velocities. Equation (1) represents the relationship between translational, tangential, and angular velocities:

$$
\begin{aligned}
v_R &= r\omega_R \\
v_L &= r\omega_L \\
v &= \frac{v_R + v_L}{2} \\
\omega &= \frac{v_R - v_L}{D}
\end{aligned} \quad (1)
$$

where, $r$ is the radius of the wheels. The robot's kinematics is defined by the coordinates $x$ and $y$ and the $\theta$ angle (the rotation of the robot relative to the base coordinate) as indicated in (2).

$$
\begin{aligned}
\dot{\theta}(t) &= \omega(t) \\
\dot{x}(t) &= v(t) \cdot \cos\theta(t) \\
\dot{y}(t) &= v(t) \cdot \sin\theta(t)
\end{aligned} \quad (2)
$$

***Assumption*** **1**- In this study, the operation workspace is assumed to be a homogeneous environment with constant friction and no slope. This study also ignores the rotational resistance/ moment of inertia of the wheels.

Based on Fig.3 (b), the dynamic model of the robot can be described by the following equations:

$$
\begin{aligned}
a(t) &= \frac{1}{m}F_L + \frac{1}{m}F_R \\
\varepsilon(t) &= \frac{-D}{2J}F_R + \frac{D}{2J}F_L
\end{aligned} \quad (3)
$$

where, $F_L$ and $F_R$ are the tangent forces applied to left and right wheels respectively by which the right and left wheels act on the road due to a change in speed wheel rotation (accelerating or braking); $m$ represents the robot's mass, $a$ is the robot's transitional acceleration, $J$ represents the inertia coefficient, and $\varepsilon$ is the robot's angular acceleration.

The Maximum force that each wheel can transmit corresponds to the frictional force $F = (F_{TL}, F_{TR})$ between the wheels and the road and can be decomposed into tangent force $F_i(i = L, R)$ and normal force $S_i(i = L, R)$, acting on the robot while cornering:

$$
\begin{aligned}
F_{T_i} &= \sqrt{F_i^2 + S_i^2}, \ i = L, R \\
S_i &= \frac{m}{2} v\omega
\end{aligned} \quad (4)
$$

Given the mathematical relations in (3), the control inputs of left and right wheels ($U_L$ and $U_R$) are calculated as follows:

$$
\begin{aligned}
J\varepsilon_L(t) + F\omega_L(t) + F_L r &= U_L \\
J\varepsilon_R(t) + F\omega_R(t) + F_R r &= U_R
\end{aligned} \quad (5)
$$

where $\varepsilon_L, \varepsilon_R$ are the angular acceleration of the left and right wheels, respectively.

Given (3) and (5), the state vector $\mathbf{x} = [\,v\ \omega\ \omega_L\ \omega_R\,]^T$ and its dynamic as $\dot{\mathbf{x}} = [\,a\ \varepsilon\ \varepsilon_L\ \varepsilon_R\,]^T$, the control vector $\mathbf{u} = [\,F_L\ F_R\ U_L\ U_R\,]^T$, and output vector $\mathbf{y} = [\omega_L\ \omega_R]^T$ are defined to develop a state space representation for the Kian-I robot as given in (6):

$$
\dot{\mathbf{x}} = \begin{bmatrix} 0 & 0 & 0 & 0 \\ 0 & 0 & 0 & 0 \\ 0 & 0 & -\frac{F}{J} & 0 \\ 0 & 0 & 0 & -\frac{F}{J} \end{bmatrix}\mathbf{x} + \begin{bmatrix} \frac{1}{m} & \frac{1}{m} & 0 & 0 \\ \frac{-D}{2J} & \frac{D}{2J} & 0 & 0 \\ -\frac{r}{J} & 0 & \frac{1}{J} & 0 \\ 0 & -\frac{r}{J} & 0 & \frac{1}{J} \end{bmatrix}\mathbf{u}
$$

$$
\mathbf{y} = \begin{bmatrix} 0 & 0 & 1 & 0 \\ 0 & 0 & 0 & 1 \end{bmatrix}\mathbf{x}
$$

(6)

## 3. System Architecture

This section presents development of a motion control architecture for a two-wheeled mobile robot. This includes two main modules of path recommender system and trajectory tracking controller. The operation diagram of the proposed control architecture is shown in Fig. 4.

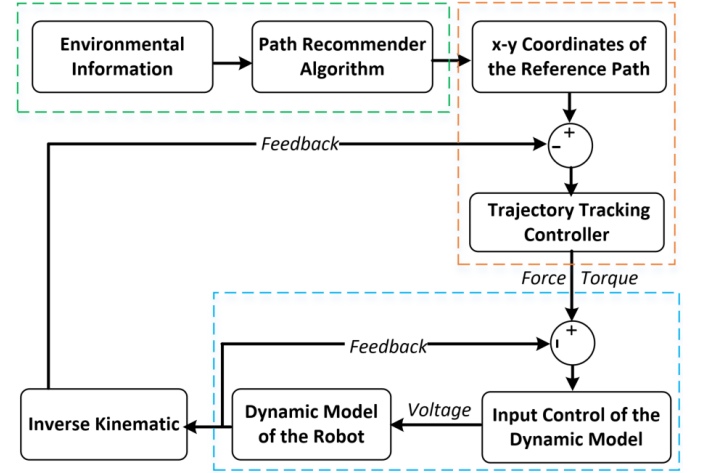

Fig 4. Control architecture of the Kian-I.

The path recommender system receives the environmental information, including the size of the playground, the positions of start and the target point, and size, shape, and location of the obstacles. Using the mentioned information, a collision-free path curve from the start point to the target point through fixed obstacles is produced. As shown in Fig.4, there are two controllers in this model. The first one is the trajectory tracking controller that receives the position error and generates force/torque for the robot, while the second controller is the input distributer that convert force/torque to the voltage and distribute the produced voltage between the wheels. In other words, the path is applied to the trajectory tracking controller, and the angular velocity of the wheels is controlled by a closed-loop feedback structure. In the subsequent subsections, details of the path recommender system and the trajectory tracking controller are provided.

## 3.1 Path Recommender System: Problem Definition and Workspace Setup

The path planning is an NP-hard optimization problem in which the main goal is to minimize the travel distance, avoiding collision with obstacle(s), and coping with environmental complexities over the time. The spline mechanism is used for determining a path curve from start point to target position passing through the obstacles. To validate performance of the path recommender system against complexity of the environment, three different workspaces with changing obstacle layouts are developed as shown in Fig.5.

The path planner should generate a time optimal collision-free path $p_i$ (shortest path) between specific pairs of start and target locations through the randomly located obstacles. The resultant path should be safe and feasible. The proposed path recommender system in this study generates potential paths $p_i:\{p_1, p_2,...\}$ using Spline curves captured from a set of control points like $\vartheta = \{\vartheta_1,...,\vartheta_i,...,\vartheta_n\}$ in the problem space with coordinates of $\vartheta_1:(\vartheta_{x(1)},\vartheta_{y(1)}),...,\vartheta_n:(\vartheta_{x(n)},\vartheta_{y(n)})$, where $n$ is the number of corresponding control points. These control points play a substantial role in determining the optimal path. The mathematical description of the B-Spline coordinates in a 2-dimensional workspace is given by:

$$\begin{cases} X(t) = \sum_{i=1}^{n} \vartheta_{x(i)} B_{i,\mathcal{K}}(t) \\ Y(t) = \sum_{i=1}^{n} \vartheta_{y(i)} B_{i,\mathcal{K}}(t) \end{cases} \quad (7)$$

$$p = [X(t), Y(t)] \approx \sum_{1}^{|p|} \vartheta_{i+1} - \vartheta_i$$

where, the $X(t)$ and $Y(t)$ are the robot's positions along the path at time $t$, $B_{i,\mathcal{K}}(t)$ is the curve blending function, and $\mathcal{K}$ is the smoothness coefficient, where larger values of $\mathcal{K}$ correspond to smoother curves. All control points should be located in respective search regions constrained to the predefined bounds of $\mathfrak{B}_\vartheta^i = [\mathcal{L}_\vartheta^i, \mathcal{U}_\vartheta^i]$. If $\vartheta_i:[\vartheta_{x(i)},\vartheta_{y(i)}]$ represents one control point in the Cartesian coordinates, the lower bound $\mathcal{L}_\vartheta^i$ and the upper bound $\mathcal{U}_\vartheta^i$ of all control points at (x-y) coordinates is calculated by (8). Then, each control point $\vartheta_i$ can be generated from (9):

$$\mathfrak{B}_\vartheta^i = [\mathcal{L}_\vartheta^i, \mathcal{U}_\vartheta^i]$$
$$\mathcal{L}_{\vartheta(x)} = [\vartheta_{x(0)}, \vartheta_{x(1)},...,\vartheta_{x(i-1)},...,\vartheta_{x(n-1)}]$$
$$\mathcal{L}_{\vartheta(y)} = [\vartheta_{y(0)}, \vartheta_{y(1)},...,\vartheta_{y(i-1)},...,\vartheta_{y(n-1)}] \quad (8)$$
$$\mathcal{U}_{\vartheta(x)} = [\vartheta_{x(1)}, \vartheta_{x(2)},...,\vartheta_{x(i)},...,\vartheta_{x(n)}]$$
$$\mathcal{U}_{\vartheta(y)} = [\vartheta_{y(1)}, \vartheta_{y(2)},...,\vartheta_{y(i)},...,\vartheta_{y(n)}]$$
$$\vartheta_{x(i)} = \mathcal{L}_{\vartheta(x)}^i + Rand_i^x\left(\mathcal{U}_{\vartheta(x)}^i - \mathcal{L}_{\vartheta(x)}^i\right) \quad (9)$$
$$\vartheta_{y(i)} = \mathcal{L}_{\vartheta(y)}^i + Rand_i^y\left(\mathcal{U}_{\vartheta(y)}^i - \mathcal{L}_{\vartheta(y)}^i\right)$$

*Definition* **1**- Performance of the generated path is evaluated based on overall path length, collision avoidance capability and constraining translational velocity of the robot as described in the following.

- **Path length minimization:** The main objective is to minimize the length of the path $\ell_p(x,y)$, which is given in (10):

$$\ell_p(x,y) = \sum_{x_1,y}^{|p|} \sqrt{\left(\vartheta_{x(i+1)} - \vartheta_{x(i)}\right)^2 + \left(\vartheta_{y(i+1)} - \vartheta_{y(i)}\right)^2} \quad (10)$$

The resultant path should be safe and feasible to environmental constraints and the robot kinematic restrictions. Two main constraints of collision avoidance and translational velocity limit are considered in this study and a penalty function is defined in (15) to form the ultimate nonlinear constrained optimization problem.

- **Collision Avoidance:** the planned path should avoid colliding obstacles regardless of complexity of environment. Therefore, a penalty function is defined to determine the violation of the collision avoidance. To this end, the distance of $i^{th}$ point of the path from the center of the $j^{th}$ obstacle should be measured, as formulated in (11).

$$d_{ij} = \sqrt{(x_i - x_{obs_j})^2 + (y_i - y_{obs_j})^2} \quad (11)$$

Where, $d_{ij}$ is the distance from $i^{th}$ point on the path to the center of the $j^{th}$ obstacle that is located in $(x_{obs_j}, y_{obs_j})$. In a case the measured distance is less than the safe boundary (radius of the obstacle) the $i^{th}$ point of the path violated the collision avoidance constraint and collision is happened. Figure 6 shows the distances of curve's points from the center of $j^{th}$ obstacle. The collision penalty function ($\mathcal{C}_{ij}$) for each arbitrary point $i$ on the path is defined by (12):

$$\mathcal{C}_{ij} = max\left(1 - \frac{d_{ij}}{r_{obs_j}}, 0\right) \quad (12)$$

In order to calculate the accumulated violation of the path from collision with a particular obstacle, $\mathcal{O}_j$ is defined to capture the mean value of all violations.

$$\mathcal{O}_j = mean([\mathcal{C}_j])$$
$$\mathcal{O}_p = \sum_{j=1}^{n} \mathcal{O}_j \quad (13)$$

where $\mho_j$ is collision violation with $j^{th}$ obstacle, $[\mathcal{C}_j]$ is the vector which includes the violation of all points for $j^{th}$ obstacle. This procedure repeats for all $n$ obstacles to obtain the overall collision violation of $\mathcal{O}_p$ for the path $p$.

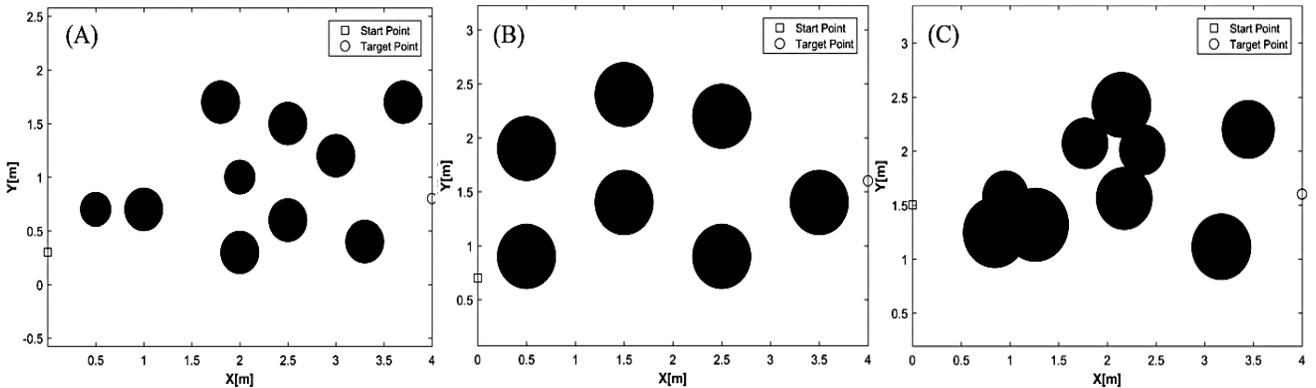

Fig. 5. Three different workspace setups for testing the performance of path recommender system.

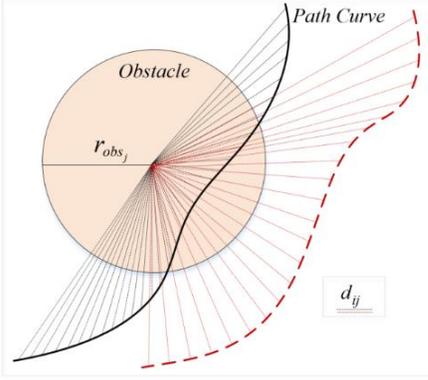

Fig. 6. Distances of random points on the path curve from the center of $j^{th}$ obstacle.

- **Violation of Transitional Velocity:** transitional velocity of the robot $(v_i)$ in position $(x_i, y_i)$ can be calculated as follows:

$$|v_i| = \sqrt{\dot{x}_i^2 + \dot{y}_i^2} \quad (14)$$

where $v_i$ is translational velocity of the robot in $i^{th}$ point, $\dot{x}_i$ and $\dot{y}_i$ are first-order derivatives of $x_i$ and $y_i$, respectively. Violation of translational velocity $(v_{\wp_i})$ in $i^{th}$ point on the path can be achieved as follows:

$$v_{\wp_i} = max\left(1 - \frac{v_{max}}{|v_i|}, 0\right) \quad (15)$$

where $v_{max}$ is the maximum velocity determined based on the kinematic of the robot. The penalty function to encounter the overall violation of the path $\wp$ is defined by (16).

$$\lambda_{v_\wp} = mean([v_{\wp_i}]) \quad (16)$$

- **Constrained Cost Function:** the path cost function is defined as a combination of the performance index, which is path length, with the associated penalty functions given in (13) and (16). Thus, a new cost function $(\hat{Z}_\wp)$ will be calculated as follows:

$$\hat{Z}_\wp = \ell_\wp \times \left(1 + \beta_\wp\left(O_\wp + \lambda_{v_\wp}\right)\right) \quad (17)$$

where $\ell_\wp$ is length of the path $\wp$, $O_\wp$ and $\lambda_{v_\wp}$ are the collision and translational velocity violations, respectively. Th $\beta_\wp$ is penalty coefficients and denote the impact of each constraint violation in calculation of total cost.

A bio-inspired optimization algorithm namely PSO is used to find a proper position of splines so that the curve passing through the control points gives the optimal (or sub-optimal) path. The argument for adopting PSO in solving NP-hard problems is strong enough due to its superior scalability with complex and multi-objective problems [20].

The PSO starts to process with initializing a population of particles, where each particle comprises a position and velocity components in the search space. The particle's position and velocity get updated iteratively using (18), and the new particles are evaluated according to the cost function given in (17), which is illustrated in Fig.7.

Each particle has a memory to preserve its experienced best position of the $\chi^{P-best}$ and the global best position of $\chi^{G-best}$ from the previous states.

$$\begin{cases} \mathcal{V}_i(t+1) = \mathcal{W}\mathcal{V}_i(t+1) \begin{array}{l} +c_1\gamma_1[\chi_i^{P-best}(t) - \chi_i(t)] \\ + c_2\gamma_2[\chi_i^{G-best}(t) - \chi_i(t)] \end{array} \\ \chi_i(t+1) = \chi_i(t) + \mathcal{V}_i(t+1) \end{cases} \quad (18)$$

Where, $\chi_i$ and $\mathcal{V}_i$ are particle position and velocity at iteration $t$; $c_1$ and $c_2$ are acceleration coefficients; $\chi_i^{P-best}$ and $\chi_i^{G-best}$ are the personal and global best positions, respectively. $\gamma_1, \gamma_2 \in [0, 1]$ are two independent random numbers; and $\mathcal{W}$ is the inertia weight and balances the algorithm between the local and global search. In each epoch, the current state value of the particle is compared with $\chi^{P-best}$ and $\chi^{G-best}$. Each particle in the swarm corresponds a potential path $\wp$, where the particle's position and velocity components are assigned with the coordinates of the Spline control points of $\vartheta_i: [\vartheta_{x(i)}, \vartheta_{y(i)}]$ that is used for path formation (refer to (7-9)). As the algorithm iterates, every particle moves toward its local best according to the outcome of the particle's individual and swarm's search. The mechanism of PSO for path recommender system is explained in Algorithm-1.

*Algorithm-1: Particle Swarm-Based Path Recommender System*
1. Assign Spline control points $\vartheta_i$ as particle position $\chi_i$
2. Initialize each particle with random $\mathcal{V}_i$ in range of $\mathfrak{B}_\vartheta^i = [\mathcal{L}_\vartheta^i, \mathcal{U}_\vartheta^i]$.
3. Choose appropriate parameters for the population size $Pop_{max}$
4. Set the number of control-points $n$ used to generate the Spline
5. Set the maximum number of iterations $Iter_{max}$
6. Initialize $\chi^{P-best}(1)$ with particle's current position at first iteration $t = 1$.
7. Set the $\chi^{G-best}(1)$ with the best particle in initial population at $t = 1$.

**For** $t = 1$ **to** $Iter_{max}$
  Evaluate each candidate particle according to given cost function
  **For** $i = 1$ **to** $Pop_{max}$
    Updated the particles $\chi_i^{P-best}$ and $\chi^{G-best}$ at iteration $t$
    **if** $\hat{Z}_\wp(\chi_i(t)) \leq \hat{Z}_\wp(\chi_i^{P-best}(t-1))$
      $\chi_i^{P-best}(t) = \chi_i(t)$
    **else**
      $\chi_i^{P-best}(t) = \chi_i^{P-best}(t-1)$
    **end (if)**
    $\chi^{G-best}(t) = \underset{1 \leq i}{argmin}\, \hat{Z}_\wp(\chi_i^{P-best}(t))$
    Update the state of the particle in the swarm
    $\begin{cases} \mathcal{V}_i(t) = \mathcal{W} \times \mathcal{V}_i(t) \begin{array}{l} +c_1\gamma_1[\chi_i^{P-best}(t-1) - \chi_i(t-1)] \\ + c_2\gamma_2[\chi_i^{G-best}(t-1) - \chi_i(t-1)] \end{array} \\ \chi_i(t) = \chi_i(t-1) + \mathcal{V}_i(t) \end{cases}$
    Evaluate each candidate particle $\chi_i$ using the given cost function $\hat{Z}_\wp(\chi_i(t))$
  **end (For)**
  Transfer best particles to next generation
**end (For)**
Output $\chi^{G-best}$ and its correlated path as the optimal solution

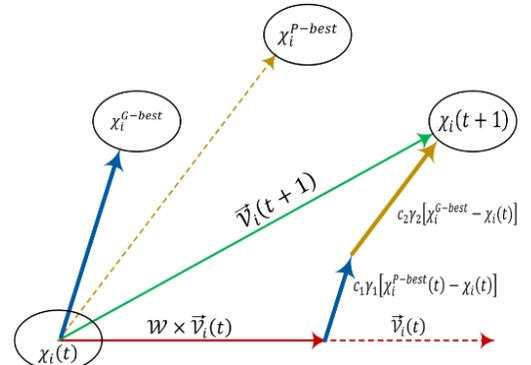

Fig.7. Process of updating the particles' position and velocity.

### 3.2 Trajectory Tracking Controller

In this research, the trajectory tracking controller receives the path information from the path planner to follow the reference path. The control design is based on PID control system which is a suitable candidate in mobile robotics [21]. A two-wheeled mobile robot moves by controlling the velocity of each wheel. The trajectory tracking controller receives error feedbacks and then the PID controllers adjust the wheels' speed to reduce the reference tracking error.

As shown in Fig.8, the control structure has two subsystems. The upper-level subsystem receives the feedback signals of $v$ and $\theta$ generated by the Inverse Kinematics block based on the $(\omega_{L_{out}},\omega_{R_{out}})$. The pair $(v,\theta)$ are compared with the reference values of $(v_{ref},\theta_{ref})$ generated based on the reference path coordinates applied in equations (19)-(20), where $\dot{x}_{ref}$ and $\dot{y}_{ref}$ are first-order derivatives of $x-y$ position components the reference path, respectively; the $(+)$ is selected for forward direction $(\dot{x} \geq 0)$, and $(-)$ is for the reverse direction $(\dot{x} < 0)$ along the $x$-axis; $k$ is a coefficient to map the positive direction of the x-axis to zero, and negative direction of the $x$-axis to 1.

$$v_{ref} = \pm\sqrt{(\dot{x}_{ref})^2 + (\dot{y}_{ref})^2} \tag{19}$$

$$\theta_{ref} = \tan^{-1}\frac{\dot{y}_{ref}}{\dot{x}_{ref}} + k\pi , \quad k = \{0,1\} \tag{20}$$

Then, the error signals $e_v$ and $e_\theta$ are fed into PID1 and PID2 to provide $\theta_{in}$ and $v_{in}$ signals to be converted to the wheels' angular velocities of $(\omega_{L_{in}},(\omega_{R_{in}})$. These signals are compared with their counterparts in the output of dynamic model $\omega_{L_{out}}$ and $\omega_{R_{out}}$ to be controlled by PID3 and PID4.

It should be noted that in Fig. 8, there are four inputs to the dynamic model which are the forces applied to the left and right wheels, as well as the voltages required to drive each wheel. Given the assumptions of the problem, the $F_L$ and $F_R$ correspond to the constant values, both are applied as step inputs to the system. Two-wheel drive voltages $U_L$ and $U_R$ are other inputs of the robot dynamic model that are generated by the PID controllers and applied to the system considering the feedback on wheel speed error received from the dynamic model. Figure 9 shows the operational diagram of PID controllers designed in MATLAB-Simulink environment.

In Fig. 9, the robot velocity $v$ and angle $\theta$ are compared with their corresponding reference values, to produce error signals for two PID controllers. Given the translational velocity of $v_{in}$ and angular velocity of $\omega = d\theta_{in}(t)/dt$, the tangential and angular velocities of each wheel can be calculated separately as follows:

$$\begin{aligned} v_L &= v + \omega \cdot D/2 \\ v_R &= v - \omega \cdot D/2 \\ \omega_L &= v_L/r \\ \omega_R &= v_R/r \end{aligned} \tag{21}$$

Using two additional PID3 and PID4 controllers, a closed-loop feedback system is constructed to determine the driving voltages of $U_L$ and $U_R$. Angular velocities from the output of dynamic model are applied to inverse kinematic block to calculate the new values of translational and angular velocities, indicated in (22).

$$\begin{aligned} v &= \frac{v_R+v_L}{2} \\ \omega &= \frac{v_R-v_L}{D} \end{aligned} \tag{22}$$

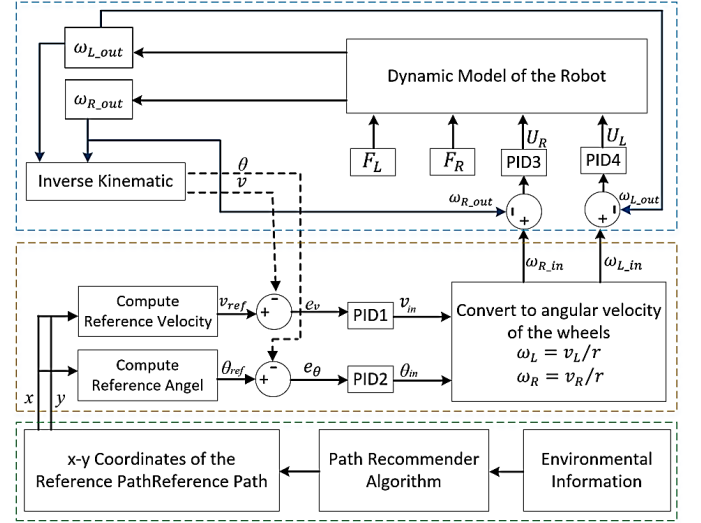

Fig.8. Block diagram of the trajectory tracking controller.

## 4. Performance Assessment and Experimental Results

To assess the feasibility of the proposed control architecture, the functionality of path recommender and trajectory controller are evaluated. The performance of path recommender system is assessed within three workspaces (depicted in Fig.5) with different barrier layouts and variable number of obstacles. The path is generated in an offline mode in the MATLAB/Simulink environment deployed on a PC computer with an Intel Core i7 processor, 2.5 GHz processing frequency, and 8 GB RAM. Then, the generated waypoints are deployed on the Arduino-Mega2560 control board of the Kian-I to assess the performance of the trajectory tracking system experimentally.

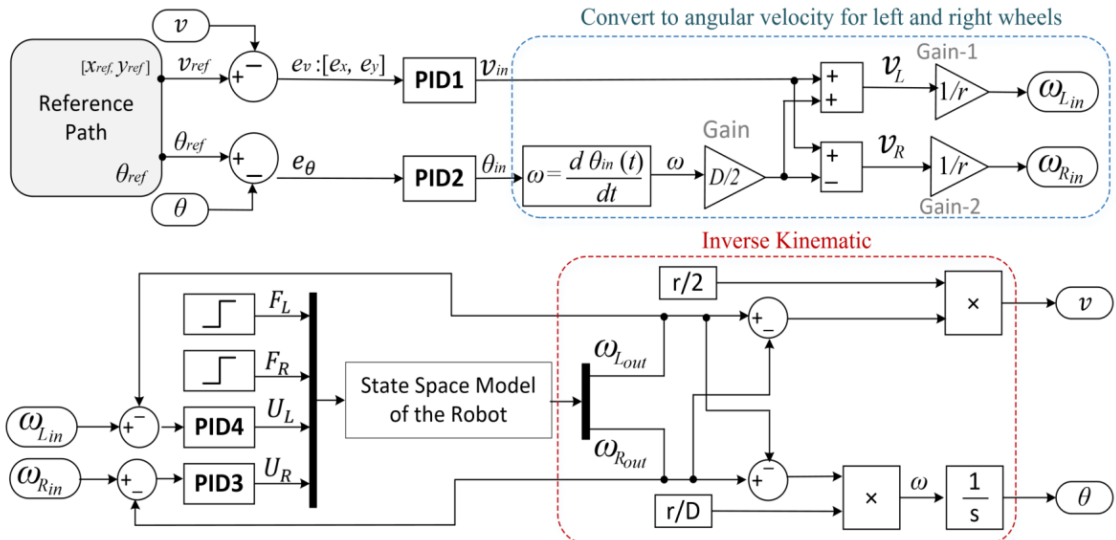

Fig. 9. Operational diagram of the tracking error control model.

### 4.1 Validation of the Kian Robot Path Recommender System

#### A. Path planning assessment in complex and unstructured environments

In this section, an extensive number of experiments are performed to evaluate the performance of the path planner. First, the performance of the planner is investigated on three different collided environments of (A), (B), and (C) (indicated in Fig.5). Then, robustness assessment of the planner based on the Monte Carlo simulations is performed. Finally, the impact of actuator constraints on the planner performance is carried out. Table 2 shows the workspace information along with the algorithm's parameter settings, determined empirically through multiple trial and error.

Table 2. PSO Parameter Settings and configuration of the workspace

| | Parameter | Value / Definition |
|---|---|---|
| **PSO Parameter Settings** | Maximum Iteration ($Iter_{max}$) | 300 |
| | Initial Population Size ($Pop_{max}$) | 100 |
| | Penalty Coefficient ($\beta_p$) | 150 |
| | Number of Splines ($\vartheta_i$) | 5 |
| | Inertia Weight ($w$) | 0.9 |
| | Personal Learning Coefficient ($c_1$) | 2 |
| | Global Learning Coefficient ($c_2$) | 2 |
| | $\sigma$ for Dtandard Deviation | 0.1 |
| **Information of the Environment** | Dimensions of search space | $4 \times 4\ m^2$ |
| | $x$-variable boundaries | $[x_{min}, x_{max}] = [0,4]$ |
| | $y$-variable boundaries | $[y_{min}, y_{max}] = [0,4]$ |
| | Radius of the obstacles | $r_{obs} = \{r_{obs_1}, \dots, r_{obs_n}\}$ $\in [0.1, 0.5]$ |
| | Center of obstacles ($x$) | $x_{obs} = \{x_{obs_1}, \dots, x_{obs_n}\}$ $\in [0.5, 3.5]$ |
| | Center of obstacles ($y$) | $y_{obs} = \{y_{obs_1}, \dots, y_{obs_n}\}$ $\in [0.5, 3.5]$ |
| | number of obstacles | $n \in [5,10]$ |
| | Start and Target position | $[x_0, y_0], [x_t, y_t]$ different for each workspace |

Figure 10 shows the shortest path generated by the path recommender system in three different workspaces, where the algorithm runs 40 times in each environment to navigate the robot from the start point (presented by yellow square) to destination point, which is depicted by green square.

In Fig. 10, the obstacles are presented in the form of black circle and have different dimensions and layouts. In Fig.10 (A) and (B), the barriers are selectively arranged, while in Fig.10 (C) obstacles are randomly scattered in the workspaces, and thus some barriers overlap and create irregular geometric shapes. The red circles along the path represent the location of spline control points, which are placed in a way to avoid colliding obstacles boundary in all three environments. The results on all three environments show that the path recommender system is successful in finding the shortest collision-free path regardless of complexity of the environment.

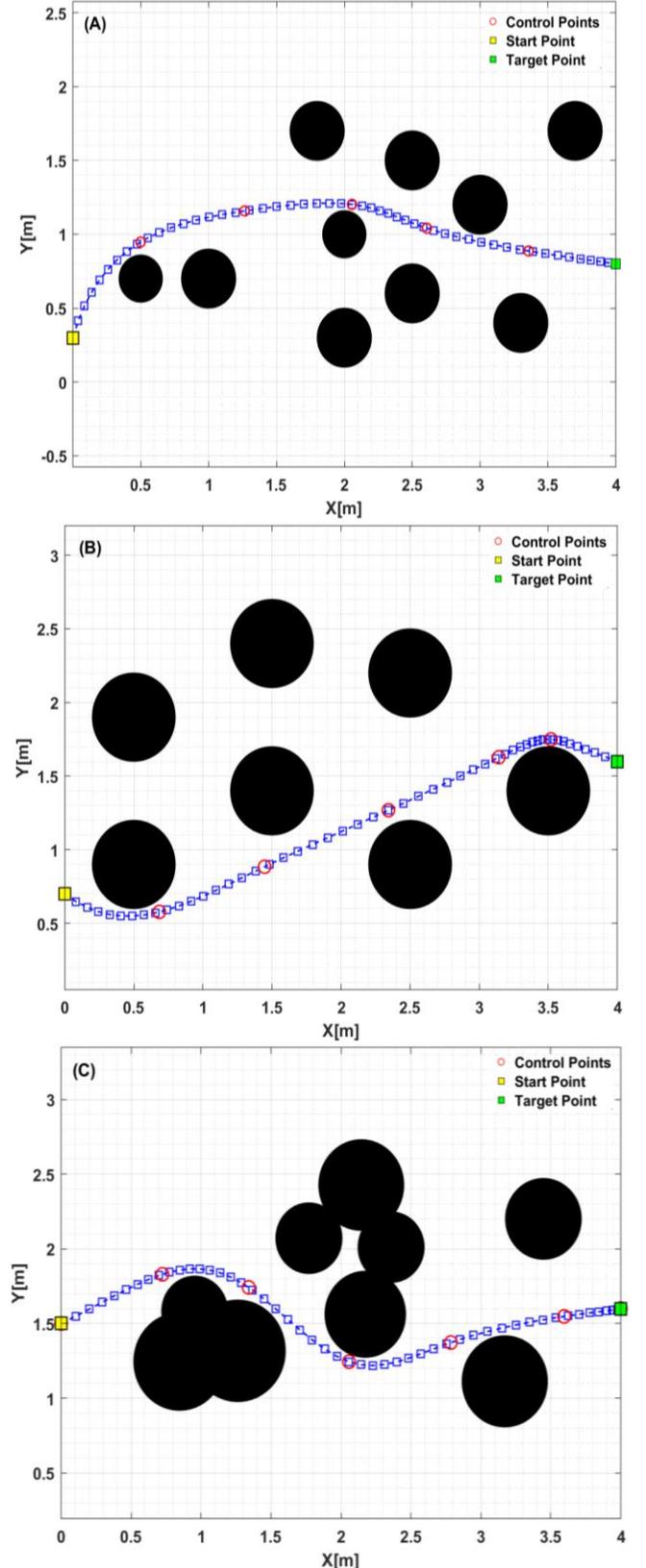

Fig.10. Shortest collision-free paths in three different workspaces of (A), (B), and (C).

To further assess the performance of the algorithm in terms of the path length, convergence rate, and computational time, statistical analysis is provided in Fig.11.

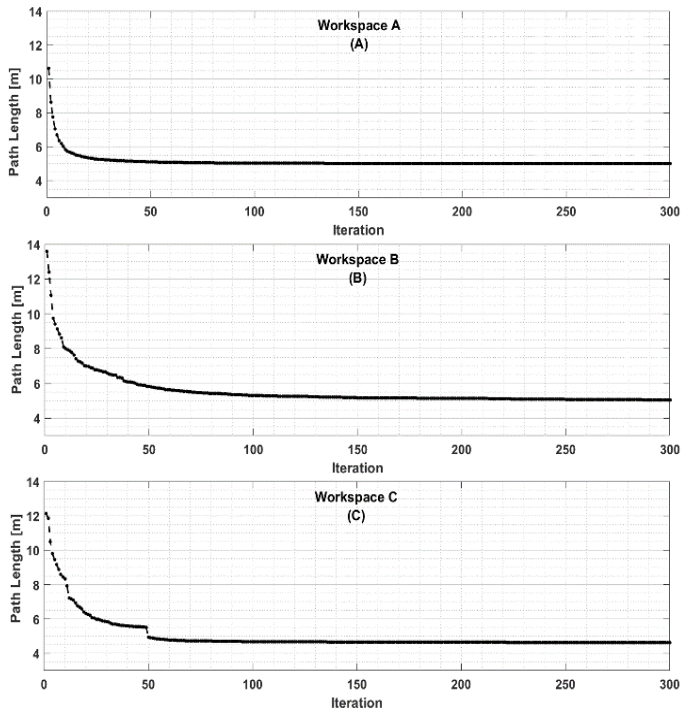

Fig.11. Optimization process of the Algorithm in terms of minimizing the path length in three test environments of A, B, and C.

As can be seen in Fig. 11, the PSO algorithm converges to its final response after approximately 30 repetitions in *A*; this occurs after approximately 100 repetitions for *B*, and 60 repetitions for *C*. This shows that complexity of the environment can impact on the convergence rate of the PSO.

*B. Robustness Analysis*

This study conducts 40 Monte Carlo simulation runs to assess the stability of the proposed model in a quantitative manner. In this test, to generate the random samples, the uniform distribution is utilized over the area of $4 \times 4$ $m^2$, where the quantity and the coordinates of the obstacles (centre and radius) are randomly generated for each run in the given area. This randomness is also applied on the robot's initial and target position. Fig.12 shows the numerical and statistical results obtained for path recommender system in environment A, B and C, while the statistical analysis carried out through 40 Monte Carlo executions. The average performance of the path recommender system based on the following metrics summarized in Fig12 (a) to (e).

− **Path Length**: The Monte Carlo executions reveals the average path length of 5.04 *m* for 40 trials on different environmental setups, which tends to be slightly longer than the average path length in A and C; however, it is notable from Fig.12 (a) that the shortest path has almost the same length for environment B and C while it is 4.5% shorter than the best produced path in A. Overall, the results of Monte Carlo simulations shows the robustness of the path recommender system against unstructured operating fields.

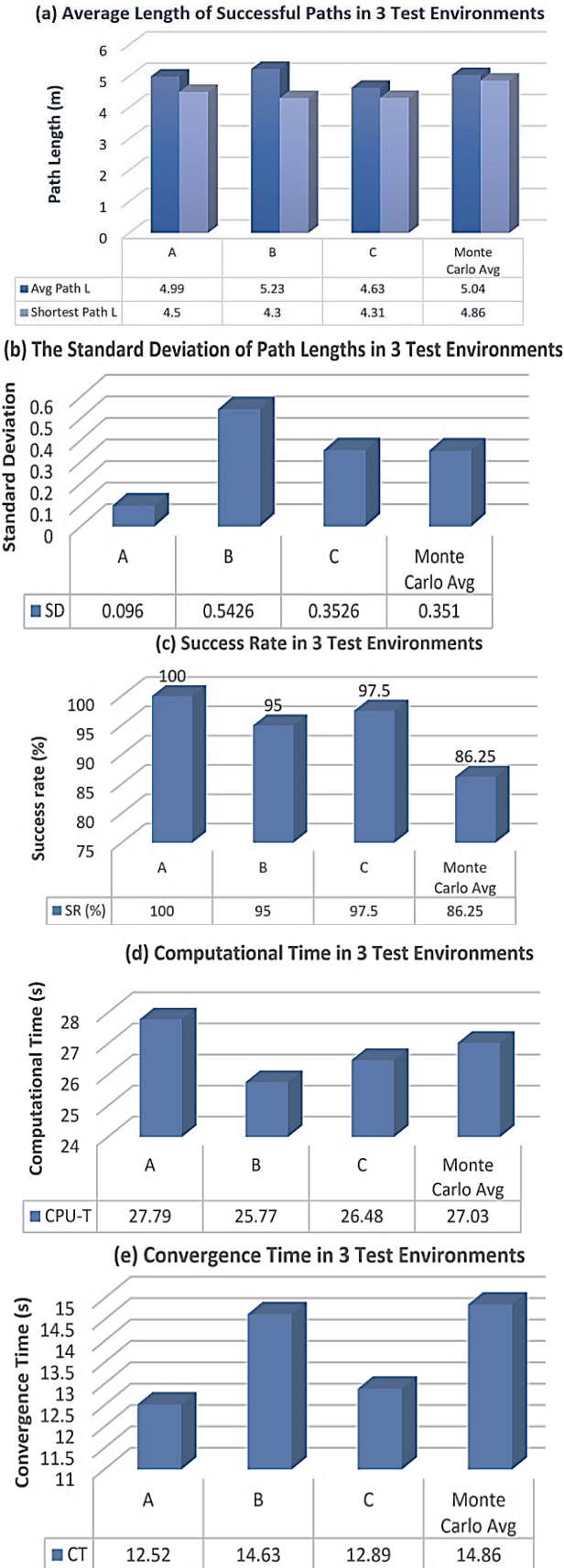

Fig.12. Performance of the path planner in three test environments of A, B and C along with the average of 40 Monte Carlo runs for: (a) Average length of successful paths and the length of the shortest path; (b) Standard deviation of the path length (SD); (c) Success rate of path planner (SR); (d) Computational time (CPU-T), (e) The algorithms convergence time (CT).

- **Standard Deviation (SD):** The Monte Carlo simulation provides an acceptable SD of the path length for 40 executions in random environments that is almost same as the SD for workspace C, 35.32% better than the SD for B. Given the small amount of Monte Carlo average SD, it is notable that the solutions for all trials are placed in a consistent range of path length, which is a good indication of performance of path recommender system.

- **Success Rate (SR):** Given the SR values in Fig12(c), the path planning in workspace A is 100% successful in finding a path without collision, while the planner system tends to have 5% and 2.5% less performance in the workspace B and C, respectively. However, considering the average SR (86.25%) and SD (0.351) values for all trials of Monte Carlo runs, the path recommender system has a high confidence in providing reasonably safe and reliable path.

- **Computation and Convergence Time:** Considering the computation time for all three environments in Fig.12 (d), the average "convergence time" can be calculated by multiplying the planner's average run time (computation time) for each environment by the mean number of iterations required for the planner to converge the optimal solution. As shown in Fig.12 (e), the planner performs considerably fast convergence with average of 14.86 sec for all Mote Carlo trials, while this time even tends to be 15.75% and 13.26% smaller for path planning in workspace A and C, respectively. It should be noted that path planning computations are performed in an interpretative programming environment of MATLAB and this can be significantly enhanced, up to the several order of magnitude, when the planning algorithm implemented by complied code of C or C++ languages.

The performance of the path recommender system is further assessed and compared by making changes to the algorithm's design parameters to investigate the impact of parameter tuning on efficiency of the PSO-based path planner. Therefore the next attempt aims at improving the solutions quality by adjusting the design parameters. The only major problem with PSO for this experiment is comparatively low success rate in workspace B. Increasing the initial population size ($Pop_{max}$) from 100 to 150 particles, reducing the inertia weight ($w$) from 0.9 to 0.7, leads a slight improvement in the performance of algorithm in workspace B as the result is presented in Fig. 13 and Table 3.

Obviously it is notable from Table.3, the provided results is improved using the new configuration as the success rate is remedied, and the standard deviation and the path length are considerably reduced. The results show that although the modifications have led to improvement in the quality of the solutions, the computational speed is reduced by 20.64 % in the new configuration due to use of larger population size. Figure 14 shows the convergence diagram of the algorithm using the new configuration.

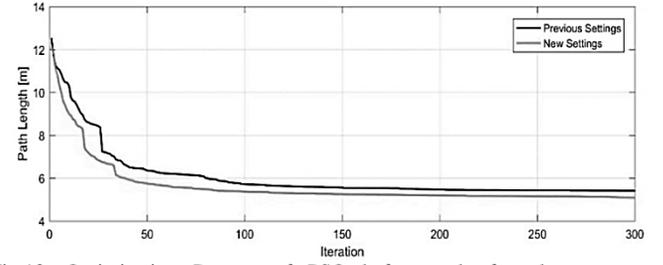

Fig.13. Optimization Process of PSO before and after the parameter adjustment

Figure 14 shows the algorithms cost convergence including best cost of population and average cost of all population members over 300 iterations using the new parameter configuration. Apparently, the best (least expensive) solution in the population is far better than the average cost of all population members, while the average cost also tends to be minimized over the 300 iterations.

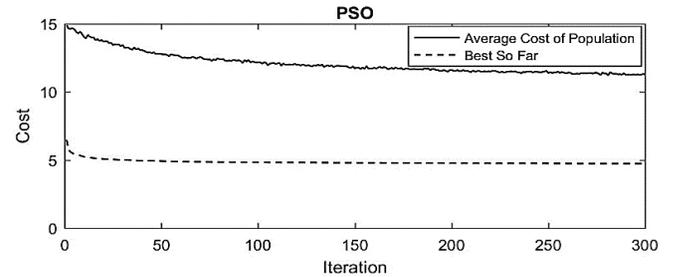

Fig.14. Convergence diagram of the algorithm using the new configuration.

In order to assess the impact of the Penalty Coefficient $\beta_p$ on the performance of the path planner, we set different values for $\beta_p$ to find out how the algorithm behaves in satisfying the defined constraints. Assigning the $\beta_p$ with a large value restricts the algorithm to strictly avoid violating the constraints, which is colliding obstacles here. The impact of changing this parameter will be examined the length of the designed paths and overall success rate of the algorithm. A benchmark on varied penalty coefficient is provided by Table 4 using the new parameter configuration, where the population size is set to be initialized with 150 particles and the inertia weight ($w$) is set to be 0.7.

It is inferable from the given results in Table 4 the increase of the penalty coefficient results in longer successful path length as the algorithm is more sensitive of avoiding crossing obstacles boundary, while the success rate of the algorithm also increased dramatically. Figure 15 shows the impact of increasing the penalty coefficient on average path length and total success rate of path planner. Setting the penalty coefficient on a small value leads to premature convergence of the algorithm and being caught in a local optima because the responses are predominantly intermittent. Therefore, increasing the value of $\beta_p$ in Fig.15 results in increment of success rate.

Table 3. Comparison on the performance of PSO in environment B, before and after parameter adjustment.

| Benchmark on Parameter Changes | Shortest Path Length (m) | Average Path Length (m) | Computation Time (s) | Success Rate (%) | Standard Deviation |
|---|---|---|---|---|---|
| Previous configuration | 4.76 m | 5.23 m | 27.13 s | 95.0 % | 0.5426 |
| New configuration | 4.65 m | 4.96 m | 34.19 s | 98.5% | 0.2644 |

Table.4. The effect of changing penalty coefficient on path planning performance

| Benchmark on varied Penalty Coefficient | $\beta_p = 50$ | $\beta_p = 100$ | $\beta_p = 150$ |
|---|---|---|---|
| Shortest Successful Path Length (m) | 4.44 | 4.48 | 4.64 |
| Average Path Length (m) | 4.98 | 5.16 | 5.35 |
| Success Rate (%) | 63.8 | 87.5 | 98.6 |

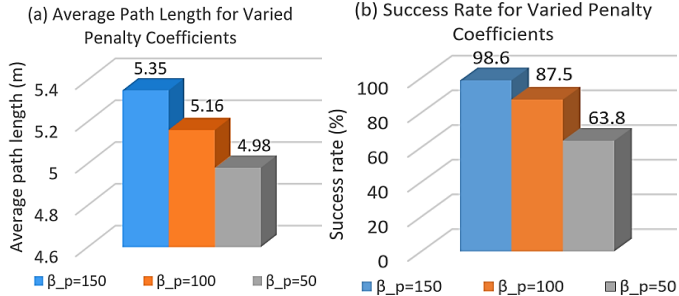

Fig. 15 Impact of changing the penalty coefficient $\boldsymbol{\beta_p}$ on the performance of the path planner for: (a) Average Path Length; (b) Success Rate.

As mentioned earlier, the designed path by the path recommender system is used as a reference path to follow by the trajectory tracking controller. Therefore, quality of the reference path has a significant impact on the performance of the trajectory tracking controller as it is obligated to follow the received reference path. The obstacle avoidance constraint is taken into account earlier and in the following section the control and dynamics constraints of the proposed model will be applied to refine the path and make it applicable on real robot (Kian-1).

### C. Planning performance assessment with input constraints

Since the robot has a restricted drive power, it cannot move at any speed and acceleration. Therefore, the applied velocity and its rate of change at the output of the track controller must not exceed a certain level. This means that, the speed and acceleration corresponding to the designed path should be compatible with the robot's dynamic and kinematic limitations to ensure a proper system performance. The inputs to the trajectory tracking will be uncontrolled and infeasible if the corresponding reference velocity and acceleration are out of range. Therefore, the rate of change of reference path components ($x_{ref}$ and $y_{ref}$) per unit of time should be considered in the path design to limit the rate of velocity and acceleration, respectively. To this end, two additional velocity and acceleration constraints are applied to the optimization problem:

$$\begin{aligned} v_{ref} - v_{max} &\leq 0 \\ a_{ref} - a_{max} &\leq 0 \\ |v_{ref}| &= \sqrt{\dot{x}_{ref} + \dot{y}_{ref}} \\ |a_{ref}| &= \sqrt{\ddot{x}_{ref} + \ddot{y}_{ref}} \end{aligned} \quad (23)$$

where, $v_{ref}$ and $a_{ref}$ are the linear velocity and acceleration corresponding to the reference path, $v_{max}$ and $a_{max}$ are the maximum operable speed and acceleration threshold (saturation level) for the robot. In order to apply the new constraints to the cost function a modified penalty function is designed in (24) and (25) that encounters all three constrains of collision avoidance, velocity, and acceleration of the reference path. The degree of violation of the new constraints on the designed path is calculated using the following penalty functions:

$$\begin{cases} v_{p_i} = \max(1 - v_{max}/|v_{ref_i}|, 0) \\ a_{p_i} = \max(1 - a_{max}/|a_{ref_i}|, 0) \end{cases} \mapsto \begin{cases} \lambda_{v_p} = mean([v_{p_i}]) \\ \lambda_{a_p} = mean([a_{p_i}]) \end{cases} \quad (24)$$

$$\hat{Z}_p = \ell_p \times \left(1 + \beta_p \left(O_p + \lambda_{v_p} + \lambda_{a_p}\right)\right) \quad (25)$$

where $v_{p_i}$ and $a_{s_i}$ are the violations associated with the velocity and acceleration with respect to the the threshold of $v_{max}$ and $a_{max}$. $\lambda_{v_p}$ and $\lambda_{a_p}$ are the average violations along the path. Figures 16 and 17 show how the new constraints affect the path design, where the maximum permitted velocity and acceleration are set on $0 \cdot 2 \ m/s$ and $0 \cdot 02 \ m/s^2$, respectively.

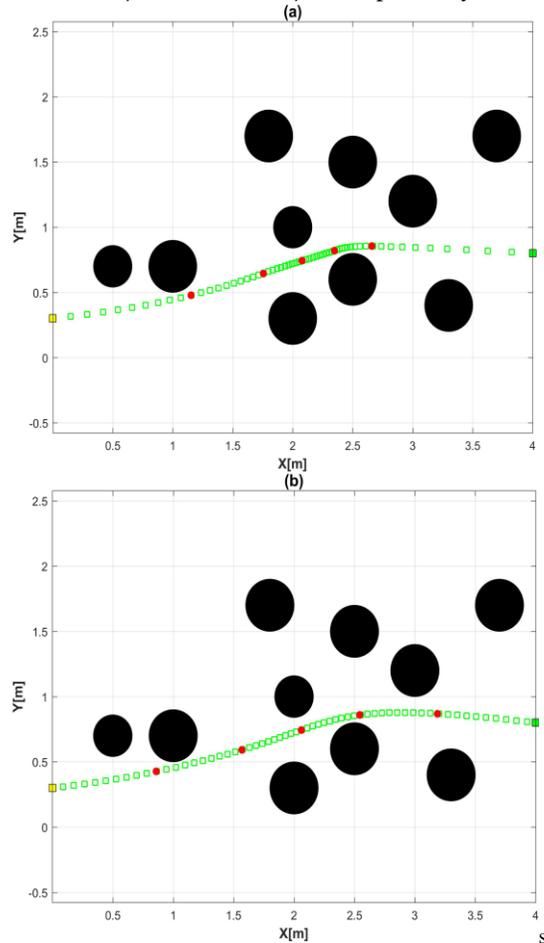

Fig.16. Influence of velocity and acceleration constraints on the path; (a): no constraints (b): with constraints.

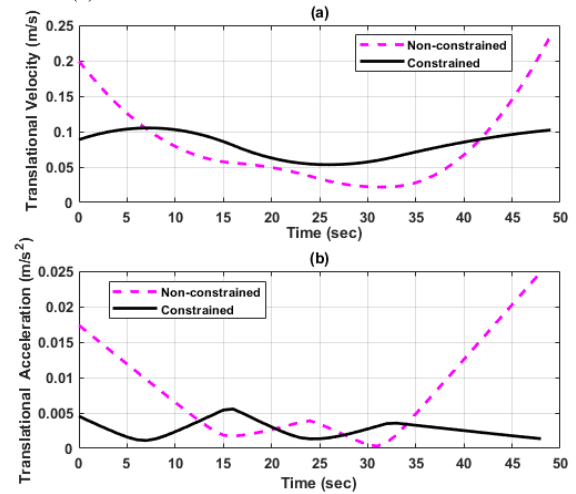

Fig.17 (a) Translational velocity before and after constraints; (b) Translational acceleration before and after the constraint.

In Fig. 16 (b) the distribution of spline control points is made more uniform and equitable allowing the path to perform better curvature with respect to barriers' layout and results in smooth and uniform adjustment of the acceleration and velocity. The velocity and acceleration diagrams corresponding to these two paths are shown in Fig.17, where the blue line represents the acceleration and velocity before applying the constraints, and the red line shows the constrained parameters. The velocity and acceleration variations in the constraint mode are much smoother and more trackable and do not exceed the maximum threshold.

Setting the Robot's actual initial position far from the designated starting point can result in positioning error and the trajectory tracking controller would not be able to move the robot from its original position to the designated path. Therefore, a trajectory needs to be designed and fed to the robot for the distance between the robot's point of departure and the path starting point. The easiest way to lead the robot from the initial position to the start point of the path is to define a straight line between these two points and add it to the designed path. However, this may result in a sharp angle in the joining point of two paths, which can be problematic for operation of the actual robot. To address this issue, a mechanism is designed in this study enabling the robot to follow the starting point of the path at each time sampling, which means the starting point shifts forward along the path over the time. This will stabilize the number of steps toward the path, reduce the length of the path, and avoid the creation of sharp turns. Figure 18 shows how to modify the robot path in the abovementioned manner.

In Fig.18, an ellipse path is designed to lead the robot from the departure spot (center of the ellipse) toward the appointed start point along the path (black square on the ellipse). As the robot moves toward the black square, the start point of path shifts forward respectively at each time step (shown by res stars). The green marks shows the robots path correction over the time while moving toward the reference path. The marked lines indicate the line of sight from green spots to the new start point on the path at each particular time frame.

Moreover, a fluctuated path with sharp angles cause sharp jumps at the control inputs, which should be avoided. Since the spline operator is used in the path design that connects the control points applying soft curves, the path recommender system in this study did not experienced any sharp angles along the path. Next section investigates the performance and accuracy of trajectory tracking controller in following the reference path designed by path recommender system.

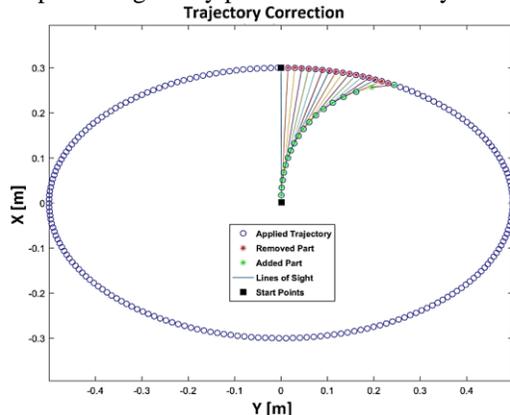

Fig.18. Mechanism of the ellipse correction with respect to the path start point

## 4.2 Evaluation of the Trajectory Tracking System and Experimental Implementation on the Kian-I Robot

The control model designed for the Kian-I robot is validated in this section. For this purpose, the generated path curves by the path recommender system in three workspaces of A, B, and C are applied to trajectory tracking controller as the reference input. The control inputs will be applied from the robot on-board processor and its performance is evaluated experimentally.

In the control model of Fig. 8, there are two feedback loops, where there are two PID controllers in the inner loop to control the speed of left and right wheels by adjusting the applied input voltage to the wheels. The outer loop also contains two other PID controllers to correct the robot's positioning error at any point in time. The system parameters including PID gain values and parameters of the model should be determined for simulation of the control structure. The PID gain values (obtained by trial and error) and the value of physical parameters of Kian-I are listed in Table 5 and 6, respectively.

Table 5. Coefficients for the PID controllers.

| Controller | Proportional coefficient ($k_p$) | Integral coefficient ($k_i$) | Derivative coefficient ($k_d$) |
|---|---|---|---|
| PID-1 | 5 | 5 | 2 |
| PID-2 | 5 | 5 | 2 |
| PID-3 | 0.5 | 5 | 2 |
| PID-4 | 0.01 | 1 | 0.1 |

Table 6. Description of the robot's dynamic model and Physical parameters of Kian-I.

| | | |
|---|---|---|
| Control Inputs ($\tau(t)$) | Forces applied to the left wheel ($\tau_1(t)$) | $F_L$ |
| | Forces applied to the right wheel ($\tau_2(t)$) | $F_R$ |
| | Voltage of left wheel ($\tau_3(t)$) | $U_L$ |
| | Voltage of right wheel ($\tau_4(t)$) | $U_R$ |
| Control Outputs ($q_{3,4}(t)$) | Angular velocity of left wheel ($\dot{q}_3(t)$) | $\omega_L(t)$ |
| | Angular velocity of right wheel ($\dot{q}_4(t)$) | $\omega_R(t)$ |
| State Variables ($q_i(t)$) | Translational velocity of robot ($\dot{q}_1(t)$) | $v(t)$ |
| | Angular velocity of robot ($\dot{q}_2(t)$) | $\omega(t)$ |
| | Angular velocity of left wheel ($\dot{q}_3(t)$) | $\omega_L(t)$ |
| | Angular velocity of right wheel ($\dot{q}_4(t)$) | $\omega_R(t)$ |
| Robot Parameters | Mass of the robot ($m$) | 0.9 $kg$ |
| | Moment of inertia ($J$) | 0.001 $kg.m^2$ |
| | Frictional force ($F$) | 0.01 $N$ |
| | Radius of the wheels ($r$) | 0.021 $m$ |
| | Distance between wheels ($D$) | 0.145 $m$ |

In the following, number of experiments are conducted to evaluate the robot's performance in two simulated and real-world environments. The trajectory tracking system is designed and verified in MATLAB/Simulink and then is deployed on the robot's on-board processor. To enhance the computational efficiency on-board the robot, the sampling time of 2.5 seconds is used. Table 7 tabulates the settings of experimental experiments.

Figure 19 shows the generated reference path and the trajectory tracking performance for all three paths in workspace A, B, and C. Comparing the given reference paths in Fig.19 (a) and the results of trajectory tracking in Fig.19 (b) it is notable that the tracking quality is perfectly satisfied and the controller successfully follows the exact pattern of the path curve. Considering Fig. 19 (c) it is inferable that the trajectory tracking controller also successfully runs on the actual robot and accurately guide the robot along the path toward the target point with no collision. Pulse Width Modulation (PWM) method was used to apply the control voltage to the motors. The controller determines the voltages needed to start the motors according to the speeds required to follow the path. These voltages must be converted to a standard signal for the Arduino controller (considering 8-bit resolution).

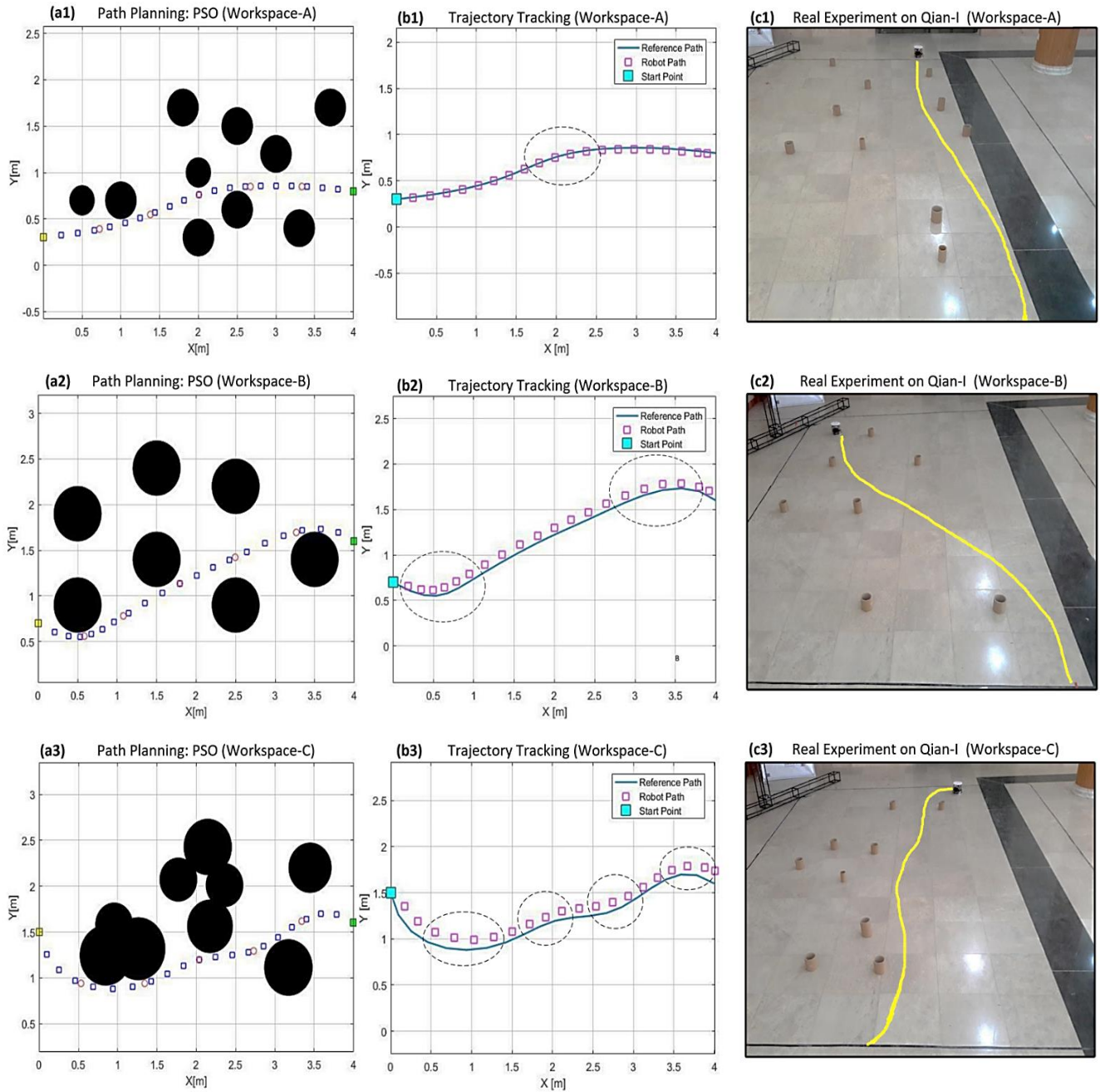

**Fig.19** (a) Designed path by path recommender system using PSO algorithm (b) Generated trajectories by the trajectory tracking controller (c) experimental results on the actual robot in the lab environment.

**Table 7.** Comparison of the experiments in simulation and actual environment.

| Parameter Setting | Simulation | Experimental |
|---|---|---|
| Number and position of obstacles | Pre-defined for each workspace | Same as the simulation set-up |
| Obstacles Radius | Variable between 0.05 to 0.2 | All in the same size with radius of 0.05 |
| Dimensions of environment | $4 \times 4\ m$ | $4 \times 4\ m$ |
| Robot Body diameter | $0.145\ m$ | $0.145\ m$ |
| Wheel radius | $0.021\ m$ | $0.021\ m$ |
| Path Time | $50\ s$ | $30\ s$ |
| Number of steps | 20 | 20 |
| Friction force | $0.01N$ | variable |

Duty cycle values of the left and right wheels for trajectory tracking in workspaces A, B, and C are shown in Fig.20. As shown in Fig.20 when the robot needs to turn right the duty cycle of the left wheel is greater than duty cycle of right wheel and vice versa. When the robot moves straight, the duty cycle values must be equal.

The trajectory tracking controller regulates the voltages applied to the left and right wheels ($U_L$,$U_R$), and used PID controller to enforce the corresponding velocity to each wheel. Numerous empirical tests have shown that if the sampling time, (2.5 seconds), is reduced to 1.5 seconds, the robot will have a better performance in tracking the reference path.

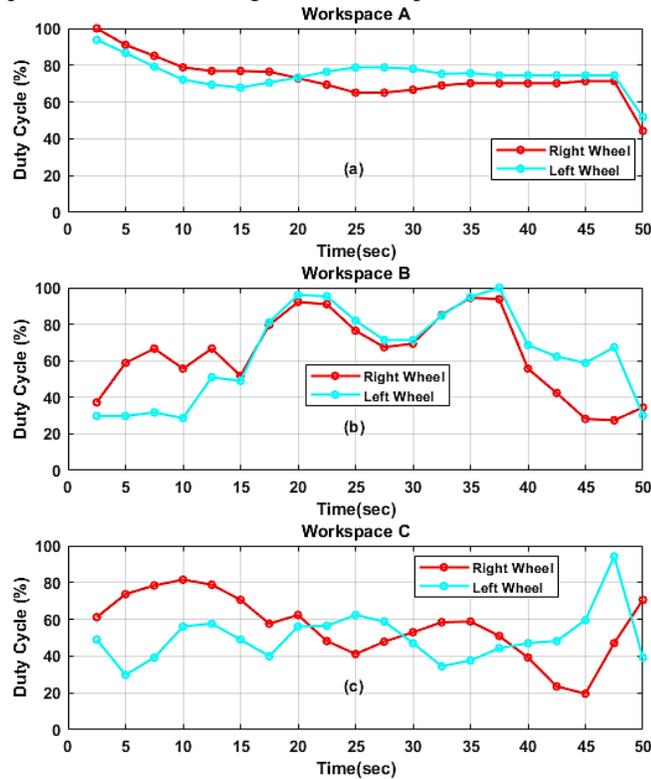

**Fig.20.** Duty cycle diagram of the wheels for the trajectory in: (a) workspace A; (b) workspace B; (c) workspace C.

Fig.21 shows the distance traveled by the robot in three environments A to C in simulation with sampling time of 2.5 (*s*) and in reality, with sampling time of 1.5 (*s*).

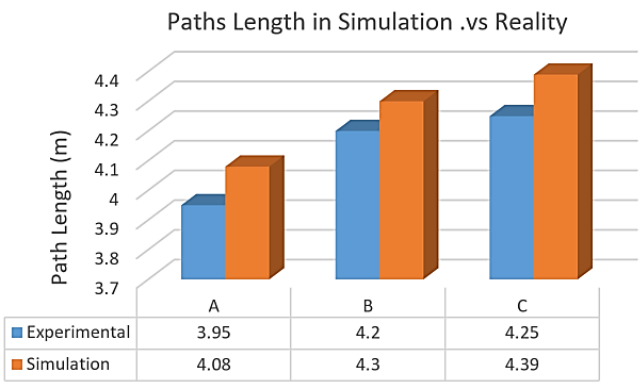

**Fig.21.** Diagram of path lengths in simulation and reality in test environments A to C.

Figure 21 declares that by adjusting the sampling time, the robot follows the designated distance with a slightly better performance, while the feasibility of the experiment with respect to robot's physical specification is assured.

## 5. Conclusions

In this paper, an affordable cost functional mobile robot called Kian-I was developed and prototyped. The robot has a similar geometrical and functional specifications similar to the Khepera-IV and is a suitable benchmark platform to verify a wide ranges of autonomous robot algorithms. More specifically, the physical design, sensor suits, and off-the-shelf on-board controller used in the prototyping phase, make the Kian-I a good candidate to be used for path planning and trajectory tracking implementation and verification. Thus, the feasibility and performance assessment of the Kian-I for path planning and trajectory tracking were investigated through both simulation and experimental results. These included performance assessment for efficient and collision-free path generation in different operating spaces cluttered by obstacles, robustness analysis via Monte Carlo tests, and assessment of fidelity and effectiveness of trajectory tracking module. The results of tests confirmed compatibility of experimental outcomes with the simulation indicating the effectiveness of the design and its suitability to be used in practice. The video of experimental tests is available in [19].

The main concentration of this paper was on the offline path planning, since the robot used the generated PWM signal to move from starting point to the endpoint. In the online path planning, the role of the sensors is more highlighted. This together with ROS integration will be included in the future works of this study to capture the online performance. In other words, the future direction for Kian-I is to be used for online collision detection, planning, and task handling in the cluttered indoor environment.